\documentclass{article}

\usepackage[final]{neurips_2019}

\usepackage[utf8]{inputenc} 
\usepackage[T1]{fontenc}    
\usepackage{hyperref}       
\usepackage{url}            
\usepackage{booktabs}       
\usepackage{amsfonts}       
\usepackage{nicefrac}       
\usepackage{microtype}      

\usepackage{graphicx}
\usepackage{algorithmic}
\usepackage{algorithm}
\usepackage{amsmath}

\title{Stochastic Channel-Based Federated Learning for Medical Data Privacy Preserving}

\author{%
  Rulin Shao\\
  Department of Mathematics and Statistics\\
  Xi'an Jiaotong University\\
  Xi'an, China \\
  \texttt{shaorulin@stu.xjtu.edu.cn} \\
   \And
   Hongyu He \\
   Department of Electrical Engineering \\
   Xi'an Jiaotong University\\
   Xi'an, China \\
   \texttt{hhy1997@stu.xjtu.edu.cn} \\
   \AND
   Hui Liu \\
   Department of Mathematics \\
   Mianyang Vocational College\\
   Mianyang, China \\
   \texttt{kaiyuanmifen@gmail.com} \\
   \And
   Dianbo Liu \\
   CSAIL \\
   MIT \\
   Cambridge, MA, USA\\
   \texttt{dianbo@mit.edu} \\
}

\begin{document}

\maketitle

\begin{abstract}
Artificial neural network has achieved unprecedented success in the medical domain. This success depends on the availability of massive and representative datasets. However, data collection is often prevented by privacy concerns and people want to take control over their sensitive information during both training and using processes. To address this problem, we propose a privacy-preserving method for the distributed system, Stochastic Channel-Based Federated Learning (SCBF), which enables the participants to train a high-performance model cooperatively without sharing their inputs. Specifically, we design, implement and evaluate a channel-based update algorithm for the central server in a distributed system, which selects the channels with regard to the most active features in a training loop and uploads them as learned information from local datasets. A pruning process is applied to the algorithm based on the validation set, which serves as a model accelerator. In the experiment, our model presents better performances and higher saturating speed than the Federated Averaging method which reveals all the parameters of local models to the server when updating. We also demonstrate that the saturating rate of performance could be promoted by introducing a pruning process. And further improvement could be achieved by tuning the pruning rate. Our experiment shows that 57\% of the time is saved by the pruning process with only a reduction of 0.0047 in AUCROC performance and a reduction of 0.0068 in AUCPR.
\end{abstract}

\section{Introduction}

Conventionally, all training data are shared with the central server. Having no control over both model training and model using processes [32], the clients may have to expose their sensitive information to the server, risking leakage of privacy.

\paragraph{Federated Learning}
McMahan [14, 30] advocates the Federated SGD and Federated Averaging algorithms as feasible approaches for the federated learning of neural networks based on iterative model averaging. Opposed to protect a single data point’s contribution in learning a model [18], R. C. Geyer [17]  proposed an algorithm for client sided federated optimization. Methods like Secure Aggregation [27] have been studied to strengthen the reliability of federated learning [26].

\paragraph{Contribution of Our Work}
The Stochastic Channel-Based Federated Learning (SCBF) could address both direct and indirect privacy leakage concerns. The clients can protect their datasets in both training and predicting process. Moreover, the inverse-model attack based on the uploaded parameters could be obstructed by the stochastic nature of our upload algorithm. Our model outperforms Federated Averaging, a state-of-the-art federated learning method. And the SCBF with Pruning (SCBFwP) could speed up the saturating of performance and save executing time. Better performance could be achieved by tuning the pruning proportion.

\section{Material and Methods}

In this section, the details of SCBF and SCBFwP are demonstrated with a specific focus on the server update procedure. 

\subsection{Stochastic Channel-Based Federated Learning}

Stochastic Channel-Based Federated Learning (SCBF) is a privacy-preserving approach which seizes the most vital information from the local training results by uploading a small fraction of gradients stochastically. The intuition behind this method is that the biological neural circuit follows the Law of Use and Disuse and the strongest neurons for an individual is those constituting an active circuit in its learning process. Correspondingly, if a channel of neurons change a lot in a training loop, we can assume it be a strong neural circuit in the network, suggesting a sensitive feature in the input sets; While the neural channels with little change in one training loop should be regarded as deteriorated ones. Choosing the channels with the most substantial variation enables SCBF to only upload a small percent of the gradients in each training loop while achieving comparable accuracy to the Federated Averaging (FA) method.

To facilitate the description of the algorithm, suppose there are $N$ features as input and a $L$-layer deep neural network is conducted with $m_1,m_2,\cdots,m_L$ neurons in each layer. For convenience sake, denote $m_0=N$ as the input dimension. Denote the wight matrix as $W=[W_1,W_2, \cdots, W_L]$ and bias matrix as $B=[B_1,B_2, \cdots, B_L]$. 

The update algorithm includes five steps: 
\paragraph{Train Local Model}
The local models are trained separately on its own datasets and each model results a gradient matrix showing the change in weight matrix during each training loop. Denote the gradient matrix as $G$. 

\paragraph{Compute Channel Norms}
Considering that a channel must go through a neuron in each layer and correlate to a $L$-dimensional vector comprising the  index of these neurons, the results of channels' norm could be saved in a $L$-dimensional tensor $T$, each element of which equals a channel norm. The shape of $T$ should be:
$$T=(t_{i_1i_2\cdots i_L})_{i_1i_2\cdots i_L=1}^{m_1m_2\cdots m_L}.$$
Denote $c^{(i)}=[g_0^{(i)},g_1^{(i)},\cdots,g_L^{(i)}]$ as $i$-$th$ channel where $\vec{i}=[i_1,i_2,\cdots,i_L]$ is the index of tensor which correlates the neurons this channel goes through in each layer; The Euclidean norm of each channel is calculated by
$$n^{(i)}=\| c^{(i)} \|_2  =\sum_{j=0}^L(g_j^{(i)})^2,$$ 
and  is saved in the $L$-dimensional tensor $T$:
$$T_{i_1,i_2,\cdots,i_L}= n^{(i)}=\| c^{(i)} \|_2  =\sum_{j=0}^L(g_j^{(i)})^2.$$

\paragraph{Sort Norms}
Given a fixed upload rate $\alpha$, we could straighten the gradient tensor to a vector and sort it, computing the $\alpha$-quantile $q_{\alpha}$ as a threshold for the channel selection.

\paragraph{Process Gradients}
There are two ways to process the gradients: 

\begin{itemize}
\item{Negative Selection:} Discard the channels whose norms are below the $\alpha$-quantile and select the rest parameters for update.
\item{Positive Selection:} Select the channels whose norm is above $q_{\alpha}$ with the rest parameters set to zeros.
\end{itemize}

\paragraph{Update Server}
Upload the processed gradient matrix $\tilde{G}$ to the server and the server updates by adding gradients $\tilde{G}$ to its original weights.

\begin{figure}
\centering
\includegraphics[width=4in]{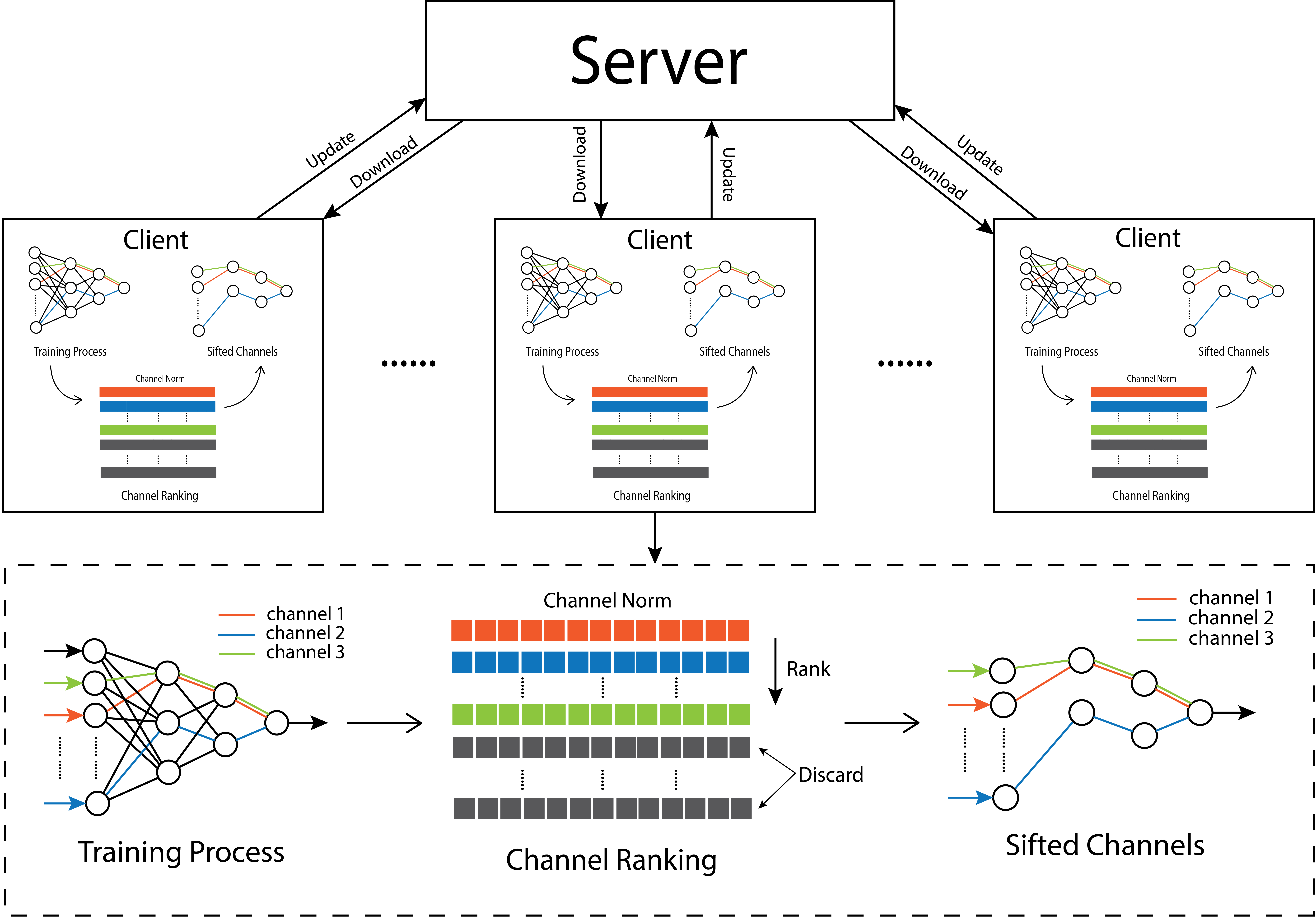}
\caption{SCBF Model}
\label{Fig: Stochastic Channel-Based Federated Learning with pruning}
\end{figure}

As shown in Fig \ref{Fig: Stochastic Channel-Based Federated Learning with pruning}, the server update algorithm is executed every global loop, and before the next training loop begins, the local model download the server's latest weights. 

\paragraph{Pruning Process}

Training a model with privacy-preserving methods could be time-consuming, especially when training sets are enormous. Addressing this problem, we introduce a neural network pruning process to SCBF. Stochastic Channel-Based Federated Learning with Pruning (SCBFwP) decides which neurons to be pruned according to APoZ [33] using validation sets.

\begin{algorithm}
\label{Alg: overall}
\caption{Pseudocode of SCBFwP}\label{update}
\begin{algorithmic}
\REQUIRE Models of local clients, model of the central server, update rate $\bf{\alpha}$, pruning rate $\bf{\theta}$, total pruned fraction $\bf{\theta_{total}}$, number of global loops, clients number $\bf{K}$
\FOR{global loops}
\FOR{each client}
\STATE Train the client model on local datasets;
\STATE Select channels according to the update rate and process the gradients $\bf{\Delta W_k}$;
\STATE Upload the processed gradients $\bf{\tilde{\Delta W_k}}$ to the server;
\ENDFOR
\STATE Update the server weights $\bf{W}$ with processed gradients from each client: 
\STATE $\bf{W} \leftarrow \bf{W} + \sum_{k=1}^{\bf{K}}{\bf{\tilde{\Delta W_k}}}$;
\IF{pruned fraction $\leq$ total pruned fraction}
\STATE Prune $\bf{\theta}$ of the server model according to validation set;
\STATE Prune each local model according to the structure of pruned server;
\ENDIF
\ENDFOR
\RETURN A distributed system with learned models
\end{algorithmic}
\end{algorithm}

\subsection{Dataset for Experiment}
Data used in our experiment was provided by hospitals, comprising 30760 admissions with status information represented by alive or expired. To explore the relationship between mortality and admissions, we develop a model that takes the medications as inputs and predictions of binary mortality as output. The cohort is managed in 2917 different medicines in total. Information on whether a patient takes each of the medicines after admission are adopted as binary input features. We use 60$\%$ of the dataset for training, 10$\%$ as the validation set, and 30$\%$ as the test set. The training set is equally divided into five parts as local training sets.

\section{Result}

\begin{figure}
\centering
\includegraphics[width=5in]{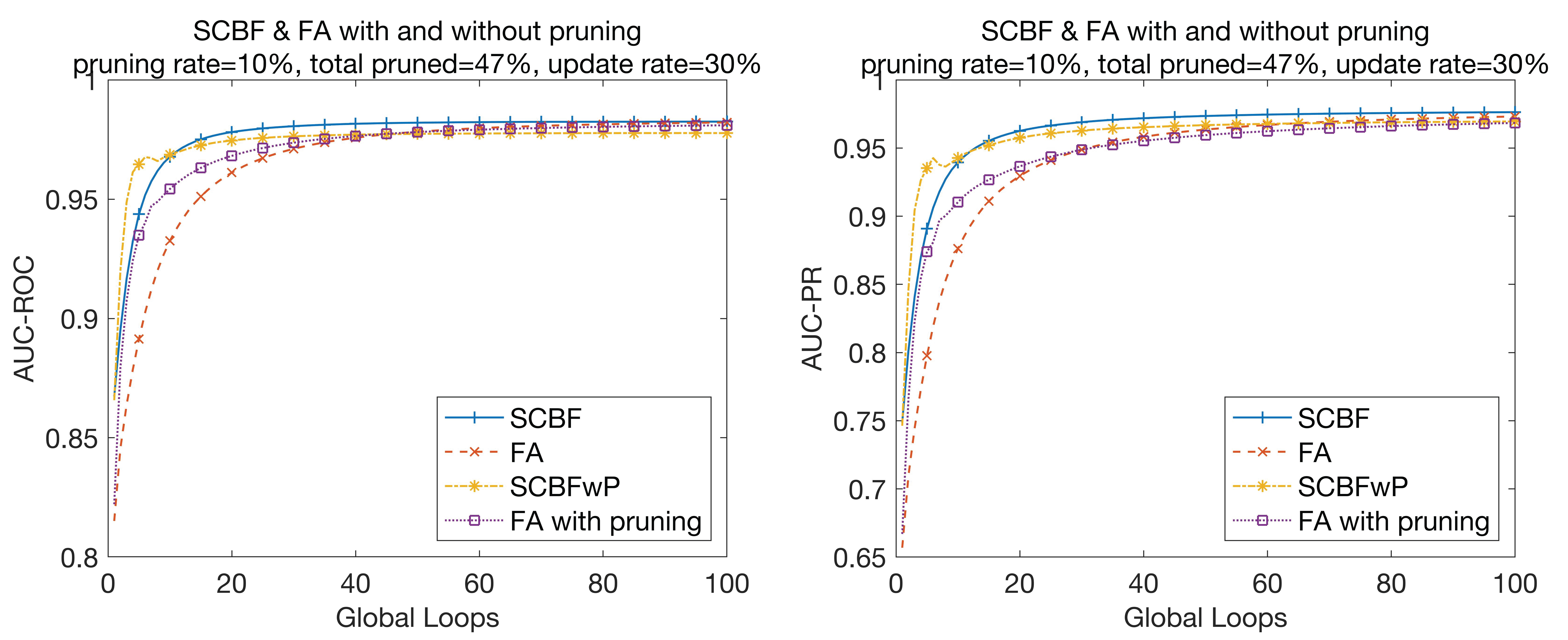}
\caption{Comparison of SCBF and FA with and without pruning. The left graph evaluate these methods by AUC-ROC, while the right one uses AUC-PR. The proposed method SCBF outperforms others by both indicators and SCBFwP method obtains the quickest saturating speed.}
\label{Fig: Vital Result}
\end{figure}

The model achieves an AUCROC of 0.9776 and an AUCPR of 0.9695 with only 10\% channels uploaded, outperforming the model which uploads all parameters. As shown in Fig \ref{Fig: Vital Result}, our model saturates much faster than the FA. The performance of SCBF keep exceeding that of FA, with half of the parameters unrevealed to the server.

To accelerate the training process and reduce the size of the neural network, we set the proportion of neurons to be pruned in one training loop (pruning rate) to 10\% and the total proportion of neurons to be pruned to 47\%. The best performance is achieved by SCBF with 0.9825 for AUCROC and 0.9763 for AUCPR, while the SCBF model with pruning outperforms others in the first 5 loops (Fig \ref{Fig: Vital Result}). Although a decline in the final performance is observed, the reduction in performances is negligible in many application situations and the trade-off between time efficiency and accuracy can be achieved by tuning the parameters of pruning process. 

Our trial shows that the SCBFwP could save 85\% of the information exchange compared to Federated Averaging. And for SCBF, the parameters uploaded to the server is 45\% by positive selection. Regarding time efficiency, pruning process could reduce 57\% of the time for SCBF and 48\% of the time for FA, releasing much burden of calculations.

\section{Conclusion}
We proposed a privacy-preserving approach for distributed systems whose models are based on any type of neural network. Our methodology develops a channel-based update algorithm to address the concern of inverse-model attacks by uploading a fraction of channels to the server stochastically, achieving a state-of-the-art performance. Moreover, we introduced a neural pruning process to the model, which could accelerate the training process and saturating speed of performances with little sacrifice of the final performances.Differential privacy could be further conducted on our models to evaluate the privacy-preserving ability quantitatively.


\section*{References}

\small

[1] Fredrikson, Matt, Somesh Jha, and Thomas Ristenpart. "Model inversion attacks that exploit confidence information and basic countermeasures." Proceedings of the 22nd ACM SIGSAC Conference on Computer and Communications Security. ACM, 2015.
\newline [2] N. R. Adam and J. C. Wortmann, Security-Control Methods for Statistical Databases: A Comparative Study, ACM Computing Surveys 21(4): 515-556 (1989).
\newline [3] R. Agrawal and R. Srikant. Privacy-preserving data mining. In Proc. ACM SIGMOD International Conference on Management of Data, pp. 439–450, 2000.
\newline [4] D. E. Denning, Secure statistical databases with random sample queries, ACM Transactions on Database Systems, 5(3):291–315, September 1980.
\newline [5] C. Dwork and K. Nissim. Privacy-preserving datamining on vertically partitioned databases. In Advances in Cryptology: Proceedings of Crypto, pages 528–544, 2004.
\newline [6] C. Dwork and A. Roth. The algorithmic foundations of differencial privacy. Foundations and Trends in Theoretical Computer Science, 9(3–4):211–407, 2014.
\newline [7] C. Dwork, F. McSherry, K. Nissim, and A. Smith. Calibrating noise to sensitivity in private data analysis. In TCC, pages 265–284. Springer, 2006.
\newline [8] Song, Shuang, Kamalika Chaudhuri, and Anand D. Sarwate. "Stochastic gradient descent with differentially private updates." 2013 IEEE Global Conference on Signal and Information Processing. IEEE, 2013.
\newline [9] Bassily, Raef, Adam Smith, and Abhradeep Thakurta. "Private empirical risk minimization: Efficient algorithms and tight error bounds." 2014 IEEE 55th Annual Symposium on Foundations of Computer Science. IEEE, 2014.
\newline [10] C. Dwork. Differential privacy. In Encyclopedia of Cryptography and Security, pages 338–340. Springer, 2011.
\newline [11] K. Chaudhuri, C. Monteleoni, and A. Sarwate. Differentially private empirical risk minimization. JMLR, 12:1069–1109, 2011.
\newline [12] A. Sarwate and K. Chaudhuri. Signal processing and machine learning with differential privacy: Algorithms and challenges for continuous data. Signal Processing Magazine, 30(5):86–94, 2013.
\newline [13] C. Dwork, G. Rothblum, and S. Vadhan. Boosting and differential privacy. In FOCS, 2010.
\newline [14] McMahan, H. Brendan, Eider Moore, Daniel Ramage, and Seth Hampson. "Communication-efficient learning of deep networks from decentralized data." arXiv preprint arXiv:1602.05629 (2016).
\newline [15] Dean, Jeffrey, Greg Corrado, Rajat Monga, Kai Chen, Matthieu Devin, Mark Mao, Andrew Senior et al. "Large scale distributed deep networks." In Advances in neural information processing systems, pp. 1223-1231. 2012.
\newline [16] Konečný, Jakub, Brendan McMahan, and Daniel Ramage. "Federated optimization: Distributed optimization beyond the datacenter." arXiv preprint arXiv:1511.03575 (2015). 
\newline [17] Geyer, Robin C., Tassilo Klein, and Moin Nabi. "Differentially private federated learning: A client level perspective." arXiv preprint arXiv:1712.07557 (2017).
\newline [18] Abadi, Martin, Andy Chu, Ian Goodfellow, H. Brendan McMahan, Ilya Mironov, Kunal Talwar, and Li Zhang. "Deep learning with differential privacy." In Proceedings of the 2016 ACM SIGSAC Conference on Computer and Communications Security, pp. 308-318. ACM, 2016. 
\newline [19] Konečný, Jakub, H. Brendan McMahan, Daniel Ramage, and Peter Richtárik. "Federated optimization: Distributed machine learning for on-device intelligence." arXiv preprint arXiv:1610.02527 (2016).
\newline [20] Konečný, Jakub, H. Brendan McMahan, Felix X. Yu, Peter Richtárik, Ananda Theertha Suresh, and Dave Bacon. "Federated learning: Strategies for improving communication efficiency." arXiv preprint arXiv:1610.05492 (2016).
\newline [21] Reddi, Sashank J., Jakub Konečný, Peter Richtárik, Barnabás Póczós, and Alex Smola. "Aide: Fast and communication efficient distributed optimization." arXiv preprint arXiv:1608.06879 (2016).
\newline [22] Ma, Chenxin, Jakub Konečný, Martin Jaggi, Virginia Smith, Michael I. Jordan, Peter Richtárik, and Martin Takáč. "Distributed optimization with arbitrary local solvers." optimization Methods and Software 32, no. 4 (2017): 813-848.
\newline [23] Shamir, Ohad, Nati Srebro, and Tong Zhang. "Communication-efficient distributed optimization using an approximate newton-type method." International conference on machine learning. 2014. 
\newline [24] Zhang, Yuchen, and Xiao Lin. "Disco: Distributed optimization for self-concordant empirical loss." International conference on machine learning. 2015.
\newline [25] Chilimbi, Trishul, Yutaka Suzue, Johnson Apacible, and Karthik Kalyanaraman. "Project adam: Building an efficient and scalable deep learning training system." In 11th {USENIX} Symposium on Operating Systems Design and Implementation ({OSDI} 14), pp. 571-582. 2014.
\newline [26] Bonawitz, Keith, Hubert Eichner, Wolfgang Grieskamp, Dzmitry Huba, Alex Ingerman, Vladimir Ivanov, Chloe Kiddon et al. "Towards federated learning at scale: System design." arXiv preprint arXiv:1902.01046 (2019).
\newline [27] Bonawitz, Keith, Vladimir Ivanov, Ben Kreuter, Antonio Marcedone, H. Brendan McMahan, Sarvar Patel, Daniel Ramage, Aaron Segal, and Karn Seth. "Practical secure aggregation for privacy-preserving machine learning." In Proceedings of the 2017 ACM SIGSAC Conference on Computer and Communications Security, pp. 1175-1191. ACM, 2017.
\newline [28] Hard, Andrew, Kanishka Rao, Rajiv Mathews, Françoise Beaufays, Sean Augenstein, Hubert Eichner, Chloé Kiddon, and Daniel Ramage. "Federated learning for mobile keyboard prediction." arXiv preprint arXiv:1811.03604 (2018).
\newline [29] Yang, Timothy, Galen Andrew, Hubert Eichner, Haicheng Sun, Wei Li, Nicholas Kong, Daniel Ramage, and Françoise Beaufays. "Applied federated learning: Improving google keyboard query suggestions." arXiv preprint arXiv:1812.02903 (2018).
\newline [30] McMahan, H. Brendan, Eider Moore, Daniel Ramage, and Blaise Aguera y Arcas. "Federated learning of deep networks using model averaging." (2016).
\newline [31] Bagdasaryan, Eugene, Andreas Veit, Yiqing Hua, Deborah Estrin, and Vitaly Shmatikov. "How to backdoor federated learning." arXiv preprint arXiv:1807.00459 (2018).
\newline [32] Shokri, Reza, and Vitaly Shmatikov. "Privacy-preserving deep learning." In Proceedings of the 22nd ACM SIGSAC conference on computer and communications security, pp. 1310-1321. ACM, 2015.
\newline [33] Hu, Hengyuan, Rui Peng, Yu-Wing Tai, and Chi-Keung Tang. "Network trimming: A data-driven neuron pruning approach towards efficient deep architectures." arXiv preprint arXiv:1607.03250 (2016).
\newline [34] Yang, Tien-Ju, Yu-Hsin Chen, and Vivienne Sze. "Designing energy-efficient convolutional neural networks using energy-aware pruning." Proceedings of the IEEE Conference on Computer Vision and Pattern Recognition. 2017.

\end{document}